\documentclass[journal]{IEEEtran}
\usepackage{amsmath,amssymb}
\usepackage{graphicx}
\usepackage{booktabs}
\usepackage{multirow}
\usepackage{array}
\usepackage{algorithm}
\usepackage{algpseudocode}
\usepackage[hidelinks]{hyperref}
\usepackage{xspace}
\usepackage{url}
\usepackage{xcolor}
\usepackage[normalem]{ulem}

\usepackage{adjustbox}
\usepackage{pifont}
\usepackage{microtype}
\newcommand{\includegraphicsmaybe}[2][]{
  \IfFileExists{#2}{\includegraphics[#1]{#2}}{
    \fbox{\parbox[c][0.22\textheight][c]{0.95\linewidth}{\centering\small Figure file missing: \texttt{#2}]}}
  }
}
\newcommand{\hmt}{\textsc{HMT}\xspace}

\newcommand{\memtree}{\mathcal{T}_{\mathrm{mem}}}

\newcommand{\obs}{o}
\newcommand{\act}{a}
\newcommand{\task}{q}
\newcommand{\hist}{h}
\newcommand{\sg}{g}

\begin{document}

\title{Enhancing Web Agents with a Hierarchical Memory Tree}

\author{Yunteng Tan, Zhi Gao, and Xinxiao Wu%
\thanks{The authors are with the Beijing Institute of Technology, Beijing 100081, China (e-mail: \{yunteng,gaozhibit,wuxinxiao\}@bit.edu.cn).}
\thanks{Corresponding author: Zhi Gao.}%
}


\maketitle

\begin{abstract}
Large language model-based web agents have shown strong potential in automating web interactions through advanced reasoning and instruction following. While retrieval-based memory derived from historical trajectories enables these agents to handle complex, long-horizon tasks, current methods struggle to generalize across unseen websites. We identify that this challenge arises from the flat memory structures that entangle high-level task logic with site-specific action details. This entanglement induces a \textbf{workflow mismatch} in new environments, 
where retrieved contents are conflated with current web, leading to logically inconsistent execution.
To address this, we propose Hierarchical Memory Tree (\hmt), a structured framework designed to explicitly decouple logical planning from action execution. \hmt constructs a three-level hierarchy from raw trajectories via an automated abstraction pipeline: the \textit{Intent level} maps diverse user instructions to standardized task goals; the \textit{Stage level} defines reusable semantic subgoals characterized by observable pre-conditions and post-conditions; and the \textit{Action level} stores action patterns paired with transferable semantic element descriptions. Leveraging this structure, we develop a \textbf{stage-aware inference} mechanism comprising a Planner and an Actor. By explicitly validating pre-conditions, the Planner aligns the current state with the correct logical subgoal to prevent workflow mismatch, while the Actor grounds actions by matching the stored semantic descriptions to the target page. Experimental results on Mind2Web and WebArena show that \hmt significantly outperforms flat-memory methods, particularly in cross-website and cross-domain scenarios, highlighting the necessity of structured memory for robust generalization of web agents.
\end{abstract}

\begin{IEEEkeywords}
LLM-based Web Agents; Hierarchical Memory; 
Cross-Website Generalization
\end{IEEEkeywords}

\section{Introduction}
\label{sec:intro}

\begin{figure*}[t]
  \centering
  \includegraphics[width=\textwidth]{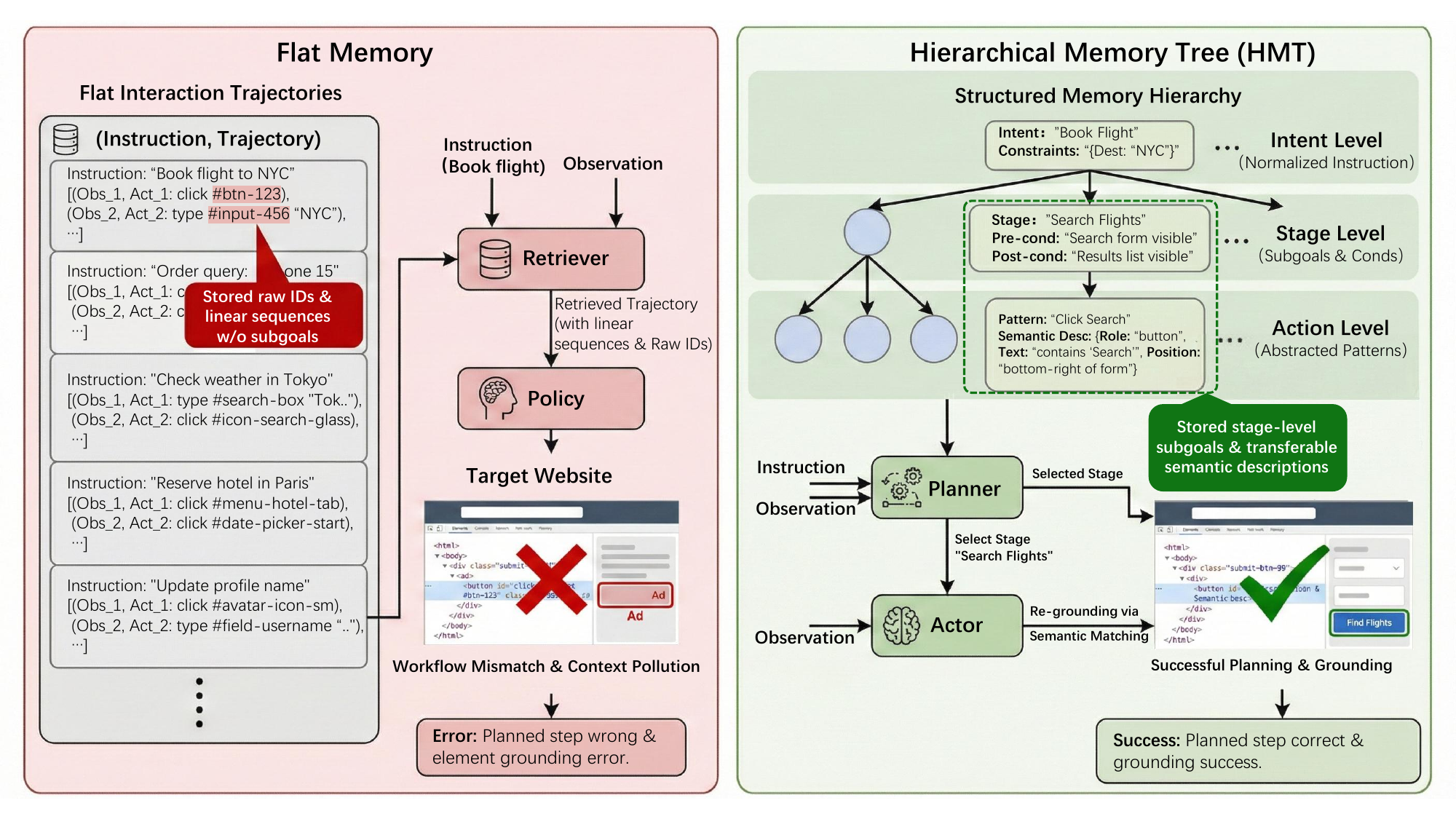}
  \caption{Comparison between Flat Memory and Hierarchical Memory Tree (\hmt). \textit{(a)} Flat memory methods retrieve interaction trajectories mixed with original workflows and source-specific implementation details w/o subgoals, leading to workflow mismatch and context pollution when applied to unseen websites. \textit{(b)} \hmt decouples intent from execution using a tree structure. It retrieves stage-aligned subgoals and abstract element descriptions, enabling the agent to plan the correct step and ground actions on the target interface effectively. }
  \label{fig:teaser}
\end{figure*}

Web agents are interactive systems designed to fulfill user-specified goals on websites by perceiving the page state and executing interface actions such as clicking, typing, and selecting. Unlike traditional automation scripts relying on fixed website structures and heuristic rules~\cite{shi2017worldbits,shi2025gui}, modern web agents leverage the reasoning and language understanding capabilities of Large Language Models (LLMs) to operate across diverse page layouts, dynamic content, and open-ended natural language instructions~\cite{yao2023react,hu2024dawn,zhang2025tongui}. This capability is instrumental for practical applications including information seeking, product comparison, service booking, and content management. In this paradigm, the agent transforms raw page observations such as textual accessibility trees or screenshots into representations processable by LLMs that subsequently predict the next optimal action.

Real-world web tasks are frequently complex and long-horizon, which require the agent to maintain the user's goal across multiple pages, handle navigation detours, and recover from local errors to ensure continued progress. To manage this complexity, handle recurrent task patterns, and learn from past experiences, such as booking round-trip flights and cross-referencing products, agents need to maintain goal consistency over extended interaction histories. Consequently, memory mechanisms have become a cornerstone of modern agent architecture. A standard design involves augmenting the agent with a retrieval-based memory that stores past successful interaction trajectories~\cite{shinn2023reflexion,wang2024agent}. By retrieving relevant trajectories based on the current instruction, the agent can utilize successful strategies as in-context demonstrations. This capability reduces redundant exploration and provides a reference for decision-making, theoretically improving efficiency for tasks with similar procedural logic.

Despite the reasoning power of LLMs, existing memory mechanisms often fail to generalize to unseen websites due to a critical bottleneck: the high-level user intents are transferable, but the specific action details (e.g., element identifiers) are not. Most existing methods store trajectories in a flat format represented as linear sequences of observations and actions. This approach entangles transferable task logic with site-specific action details. When such a trajectory is retrieved in a novel environment, it introduces an issue where the related high-level intent is paired with invalid low-level action details. For example, the agent may attempt to click a button with an ID that does not exist on the new site. We refer to this phenomenon as \emph{intention-execution entanglement}. This entanglement leads to workflow mismatch and context pollution, where the agent retrieves actions that are functionally correct for the source task but sequentially invalid on the target page (e.g., skipping necessary navigation steps), as shown in Fig.~\ref{fig:teaser}(a). This context mismatch constitutes a primary obstacle to achieving robust generalization because the memory is retrieved based on intent similarity but impedes execution due to brittle grounding details.

To address this challenge, we propose \emph{Hierarchical Memory Tree} (\hmt), a structured framework designed to explicitly decouple logical planning from action execution. \hmt constructs a three-level hierarchy from raw trajectories via an automated abstraction pipeline. First, the \emph{Intent Level} maps diverse user instructions to standardized intents to stabilize retrieval against phrasing variations. Second, the \emph{Stage Level} identifies reusable semantic subgoals characterized by observable \emph{pre-conditions} and \emph{post-conditions}. These conditions allow the agent to align retrieval with its current progress based on observable page states rather than just the initial instruction. Third, the \emph{Action Level} stores action patterns paired with transferable semantic element descriptions. These descriptions capture transferable features such as the role, label, relative position, and structural context of a target element rather than site-specific identifiers. This hierarchical design prevents invalid execution details from propagating to new environments while preserving the procedural logic necessary for task completion.

Leveraging this memory structure, we develop a stage-aware inference mechanism comprising a Planner and an Actor for inference. The Planner identifies the appropriate high-level functional stage by matching the current observation against the stored pre-conditions and post-conditions. Once the stage is identified, the Actor grounds abstract step patterns onto the current page by matching the stored semantic descriptions to candidate elements in the current page. To further enhance robustness, this process is augmented with a confidence-aware fallback mechanism to handle uncertain retrieval or stage selection (see Fig.~\ref{fig:teaser}(b)). We evaluate \hmt under two realistic protocols,  including an offline setting using Mind2Web to test transferability from pre-built memory, and an online setting using WebArena to test the accumulation of procedural knowledge during deployment. Experimental results demonstrate that \hmt effectively mitigates intention-execution entanglement, yielding consistent improvements in success rates under cross-website and cross-domain distribution shifts.

In summary, this paper makes the following contributions:
\begin{itemize}
    \item We propose \hmt, a hierarchical memory architecture for web agents, which mitigates \emph{intention-execution entanglement} by organizing stored interaction trajectories into a hierarchy of intent, stage, and action levels.
    \item We introduce a step-level memory abstraction method that stores action patterns together with transferable semantic element descriptions, enabling element grounding on new websites without relying on raw element identifiers.
    \item We develop a stage-aware inference mechanism based on top-down retrieval and a Planner-Actor decomposition for inference, augmented with a confidence-aware fallback to handle uncertain retrieval or stage selection.
    \item We empirically evaluate \hmt on Mind2Web and WebArena, demonstrating that it significantly outperforms representative flat-memory methods in cross-website settings.
\end{itemize}

\section{Related Work}
\label{sec:related}

\subsection{LLM-based Web Agents}
The development of web agents has evolved through distinct phases. Early approaches primarily relied on reinforcement learning or imitation learning trained on hand-crafted features derived from Document Object Model (DOM) trees or accessibility trees. Methods such as World of Bits~\cite{shi2017worldbits}, DOMNet~\cite{liu2018domnet}, and early DQN-based approaches~\cite{mnih2015human} utilize graph neural networks to encode element relationships. While effective in constrained environments, these methods struggle with the diversity of real-world websites due to sparse rewards and high exploration costs, a challenge also observed in mobile device control datasets like Android in the Wild~\cite{rawles2024android}. The advent of LLMs shifts the paradigm toward agents that leverage pre-trained knowledge for instruction following and reasoning~\cite{li2025efficient,yao2022webshop,wang2023survey,openai2023gpt4}.

Foundational frameworks such as ReAct~\cite{yao2023react} and Tree of Thoughts~\cite{yao2024tree} enable agents to interleave reasoning traces with action execution. Specific to the web domain, systems like WebGPT~\cite{nakano2021webgpt} demonstrate browser-assisted question answering. Following this, general-purpose agents like MindAct~\cite{deng2023mind2web} and SeeAct~\cite{zheng2024seeact} focus on grounding actions in HTML or screenshots. Nevertheless, directly processing raw HTML often leads to information overload. Recent works like Prune4Web~\cite{li2025prune4web} and HtmlRAG~\cite{zhao2024htmlrag} demonstrate that dynamically pruning the DOM tree based on current sub-tasks significantly improves grounding accuracy. This validates our design choice of using a hierarchical structure to filter context. Furthermore, research on multimodal agents, such as CogAgent~\cite{hong2024cogagent} and AppAgent~\cite{yang2024appagent}, explores visual grounding on mobile and desktop GUIs. Techniques like Set-of-Mark (SoM) prompting~\cite{yang2023som} are also proposed to overlay visual markers for better element identification. Complementing these architectures, our work introduces a structured memory backbone that leverages semantic element descriptors to actively filter observation noise, thereby addressing the grounding fragility inherent in general-purpose agents.

\subsection{Memory and Experience Reuse}
To support long-horizon autonomy, augmenting agents with external memory has become a critical research direction. Retrieval-augmented generation (RAG)~\cite{lewis2020retrieval,karpukhin2020dense} allows agents to access relevant textual knowledge~\cite{asai2024selfrag}, while episodic memory modules enable the retrieval of past interaction trajectories to guide future actions. For instance, Reflexion~\cite{shinn2023reflexion} persists verbal reinforcement to prevent repeated errors. In the context of open-ended exploration, systems like Voyager~\cite{wang2023voyager} maintain a skill library to store complex behaviors. More advanced memory architectures, such as MemGPT~\cite{packer2024memgpt}, manage context via tiered storage systems similar to operating systems. Specific to web navigation, Agent Workflow Memory (AWM)~\cite{wang2024agent} induces reusable workflows from successful trials to guide agents in both offline and online settings, and EchoTrail~\cite{pan2025echotrail} further refines this by utilizing critic-guided exploration to filter high-quality trajectories.

Typically, these retrieval-based methods employ a flat memory structure that stores trajectories as linear sequences. While effective in fixed environments, this rigid format risks entangling high-level intent with site-specific action details, leading to context pollution when applied to new environments. This challenge mirrors problems in long-term user behavior modeling and cold-start recommendation, where recent literature suggests solutions like offline trajectory clustering~\cite{zhou2025encode} and meta-learning for domain adaptation~\cite{guan2023metacdr}. Inspired by these directions, our automated abstraction pipeline can be viewed as a form of explicit, task-driven clustering of behavioral sequences. Although recent works like SkillEvo~\cite{zhang2025skillevo} and AriGraph~\cite{anokhin2025arigraph} have begun exploring graph-based methods for structuring procedural knowledge, \hmt uniquely differentiates itself by restructuring memory into a specific hierarchy designed to explicitly decouple logical planning from execution. This design effectively mitigates the context mismatch problem found in flat retrieval baselines.

\subsection{Hierarchical Planning and Abstraction}
Hierarchical structures have long been utilized to manage complexity in planning and robotics. Hierarchical reinforcement learning decomposes tasks into sub-policies with initiation and termination criteria to handle long time horizons~\cite{sutton1999between}. Similarly, decomposition-based prompting strategies guide models to solve problems via intermediate steps~\cite{zhou2022least,wang2023plan}, building upon the foundations of Chain-of-Thought (CoT) prompting~\cite{wei2022chain} and Self-Consistency~\cite{wang2022self}. More recently, advanced planning algorithms like Language Agent Tree Search (LATS)~\cite{zhou2023lats}, CRITIC~\cite{gou2024critic}, and multi-agent frameworks like AutoGen~\cite{wu2023autogen} integrate Monte Carlo Tree Search and self-verification to enhance reasoning robustness.

To support such reasoning in long-horizon contexts, recent literature explores structured memory representations. For example, MemTree~\cite{rezazadeh2025memtree} organizes dialogue history into a dynamic tree for efficient retrieval, while HiAgent~\cite{hu2025hiagent} employs hierarchical subgoal chunks to manage working memory. Such hierarchical paradigms have also proven effective in decoupling complex reasoning tasks across various domains. For instance, PR4SR~\cite{cao2025pr4sr} decouples session-level context setting from path-level reasoning in recommendation, while HSMH~\cite{wang2024hsmh} fuses local and global knowledge for multi-hop reasoning on graphs. Furthermore, for handling rare or unseen interactions, HMLS~\cite{zheng2024fewshot} demonstrates that hierarchical structures enable knowledge sharing in few-shot scenarios. Tool-use frameworks like ToolLLM~\cite{qin2024toolllm} also emphasize the importance of structured API calls. In the agent domain, LGRL~\cite{wang2025lgrl} and LDSC~\cite{shek2025ldsc} validate the ``Planner-Actor" paradigm, where a high-level Planner dictates semantic subgoals and a low-level Actor handles local execution. WebOperator~\cite{yan2025weboperator} further incorporates action-aware tree search to handle reversible actions. While these methods focus on real-time planning, they often lack a mechanism to persist successful plans for future reuse. \hmt bridges this gap by crystallizing the Planner-Actor decomposition into a permanent memory structure. Uniquely, \hmt introduces a dedicated abstraction pipeline at the leaf level, allowing the agent to not only decompose tasks logically via subgoals but also structurally decouple the action pattern from its specific realization on the target webpage.

\section{Method}
\label{sec:method}

\begin{figure*}[t]
  \centering
  \includegraphics[width=\textwidth]{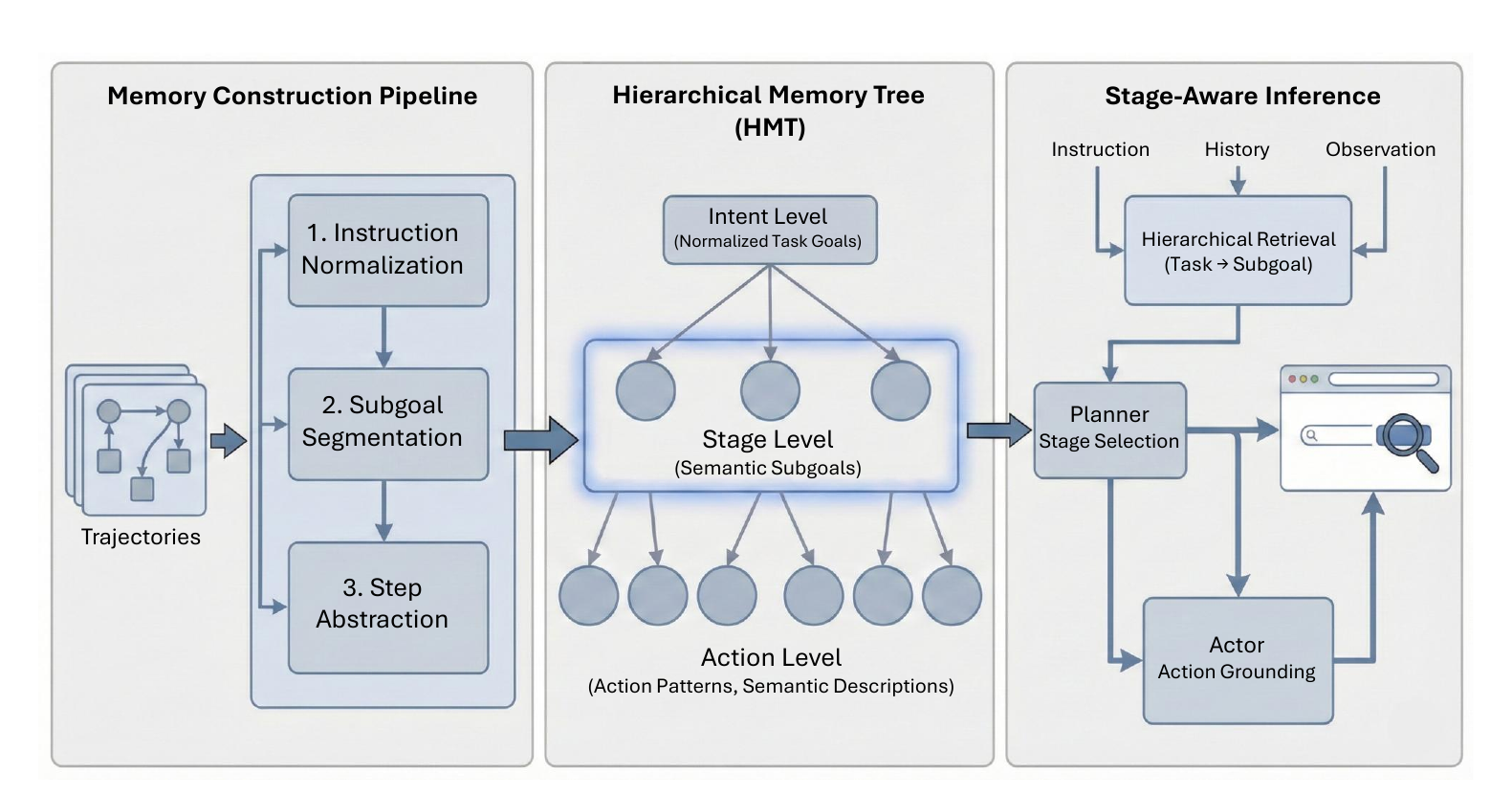}
  \caption{Overview of \hmt. The framework consists of a construction pipeline that abstracts raw trajectories into a hierarchical memory tree, and a stage-aware inference mechanism where a Planner selects the logical stage and an Actor grounds the action level description to the target page.}
  \label{fig:overview}
\end{figure*}

This section presents \hmt, a hierarchical memory framework designed to decouple reusable task logic from website-specific action details. We first formalize the web agent setting and the memory representation, and then introduce the memory construction pipeline and the stage-aware inference process. An overview of the entire framework is illustrated in Fig.~\ref{fig:overview}.

\subsection{Problem Formulation}
\label{sec:problem}

We consider a web task where an agent receives a natural-language instruction $\task$ and interacts with the browser over discrete time steps $t=1,\dots,T$. At step $t$, the agent receives an observation $\obs_t$ such as a simplified DOM tree, accessibility tree, or screenshot, and maintains the interaction history $\hist_{t-1}=\{(\obs_1,\act_1),\dots,(\obs_{t-1},\act_{t-1})\}$, and $a$ denotes actions.

Following common benchmark interfaces~\cite{deng2023mind2web,zhou2024webarena}, we assume that $\obs_t$ can be processed into a set of candidate UI elements $\mathcal{E}_t=\{e_{t,1},\dots,e_{t,K}\}$, where each UI element $e_{t,k}$ contains attributes such as visible text, role, and bounding box. An action is formally represented as a tuple $\act_t=(\mathrm{op}_t, \mathrm{target}_t, \mathrm{arg}_t)$, where $\mathrm{op}_t$ denotes the operation type like \textsc{click} or \textsc{type}, $\mathrm{target}_t$ refers to a unique identifier for an element in $\mathcal{E}_t$ such as a backend node ID, and $\mathrm{arg}_t$ is an optional argument.

To facilitate long-horizon planning, we augment the agent with a hierarchical memory tree $\memtree$. We define a retrieval function $R$ that returns a relevant memory path $\mathcal{P} = R(\task, \hist_{t-1}, \obs_t; \memtree)$. The policy is decomposed into two stages: a Planner $\pi_{\text{plan}}$ that generates a logical subgoal $\sg_t$, and an Actor $\pi_{\text{act}}$ that produces the grounded action $\act_t$:
\begin{align}
\sg_t &= \pi_{\text{plan}}(\task, \hist_{t-1}, \obs_t, \mathcal{P}), \\
\act_t &= \pi_{\text{act}}(\task, \hist_{t-1}, \obs_t, \sg_t, \mathcal{P}).
\end{align}
This separation ensures that planning operates on transferable logic, while action generation handles local grounding.

\subsection{Hierarchical Memory Tree}
\label{sec:memtree}

\begin{figure*}[t]
  \centering
  \includegraphics[width=\textwidth]{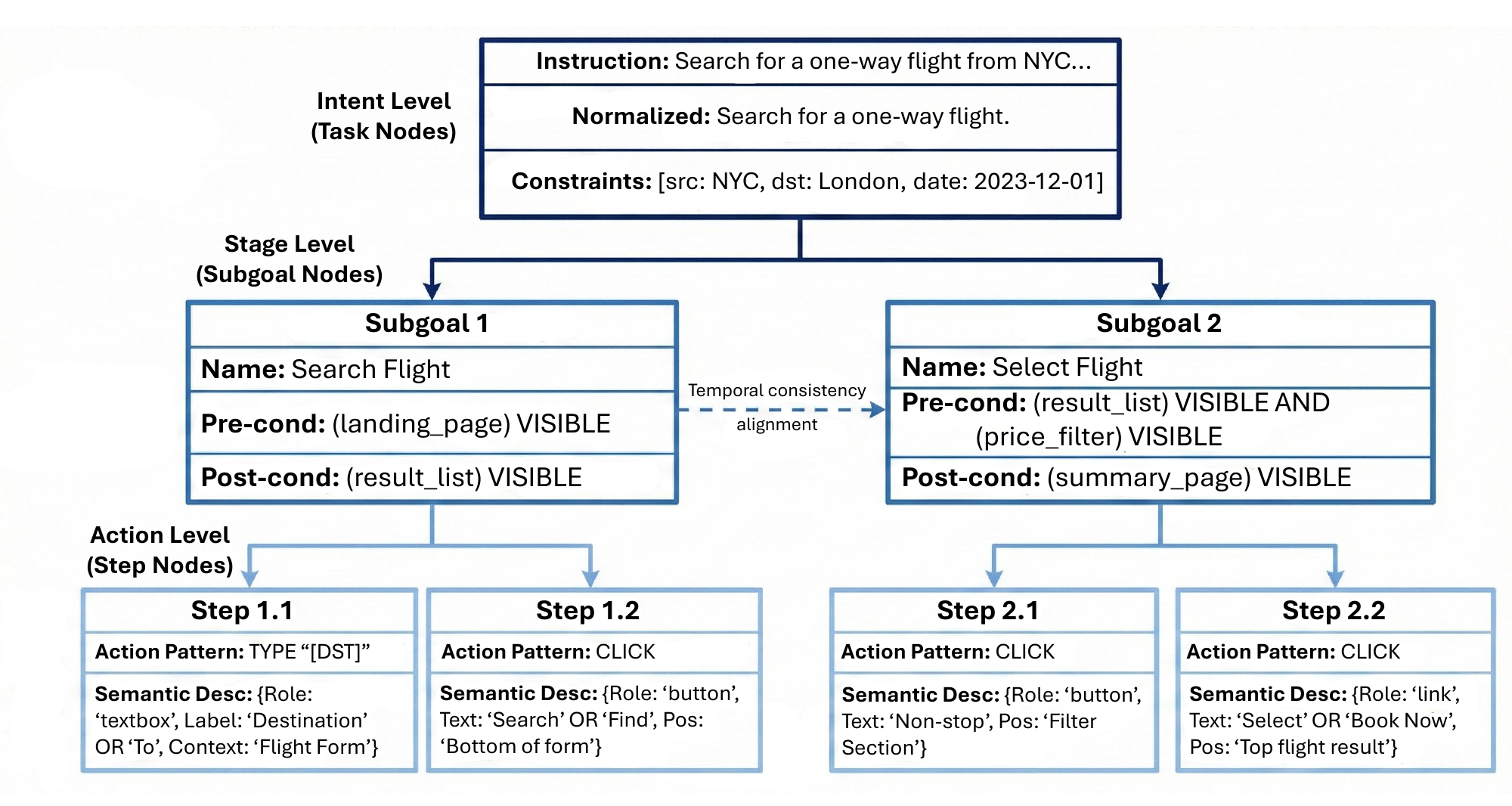}
  \caption{Structure of the Hierarchical Memory Tree. Unlike flat lists, \hmt organizes memory into intent, stage, and action levels.}
  \label{fig:tree_structure}
\end{figure*}

The architecture of \hmt is guided by three principles essential for generalization: transferability by abstracting away site-specific details, stage-awareness by aligning retrieval with execution progress, and compactness to fit within context windows. Formally, $\memtree$ is a rooted tree (see Fig.~\ref{fig:tree_structure}) comprising three levels of abstraction designed to avoid mismatch at different granularities.

At the root, the \emph{Intent Level} consists of task nodes that map diverse user instructions to standardized task goals paired with constraints. For example, a raw request like ``I want to fly to New York" is normalized to ``Intent: Book a flight; Constraints: to NYC". This standardization strips away phrasing variations from raw interaction trajectories, ensuring that diverse user queries map to the same canonical task.

At the second layer, the \emph{Stage Level} branches into reusable semantic subgoals representing functional stages of the workflow such as ``Filter Results". Crucially, to enable stage-aware retrieval, each node is characterized by explicit \emph{pre-conditions} and \emph{post-conditions} described in terms of observable UI states. For instance, a pre-condition might specify ``Search results visible", while the corresponding post-condition might state ``Price filter applied". This allows the agent to identify its current progress based on visual evidence rather than solely on potentially noisy history.

Finally, the leaf layer constitutes the \emph{Action Level}, providing executable guidance. To prevent brittle transfer caused by site-specific action details, nodes in this level exclude raw element identifiers or coordinates. Instead, they encode an \emph{action pattern} representing the semantic operation paired with a \emph{transferable semantic element description} containing attributes like role, label, and relative position. By storing directives such as ``Click the `search' button" rather than ``Click \#btn-123," \hmt ensures that the retrieved memory remains valid across different webs.

\subsection{Memory Construction}
\label{sec:construction}

We build \hmt from successful interaction trajectories $\mathcal{D}=\{(\task_i,\tau_i)\}$, where $\tau_i=\{(\obs_{i,t},\act_{i,t})\}_{t=1}^{T_i}$ represents the sequence of observations and actions. The construction process involves a unified pipeline applied under different settings.

\subsubsection{Construction Pipeline}

The pipeline begins with \emph{Instruction Normalization}, where we employ an LLM to rewrite the raw instruction $\task_i$ into a normalized intent and set of constraints. We enforce a structured JSON output format to ensure determinism and to cluster semantically identical requests into a single task node.

Following normalization, we perform \emph{Subgoal Segmentation} to partition the trajectory $\tau_i$ into contiguous segments $\{G_{i,k}\}$ using an LLM. We generate semantic names and pre-conditions/post-conditions for each subgoal, enforcing strict temporal consistency checks including coverage, ordering, contiguity, and minimum length to ensure structural validity. If any check fails, a fallback mechanism triggers a \emph{coarser segmentation} by iteratively merging adjacent problematic segments until structural integrity is restored.

Once the hierarchy is established, we proceed to \emph{Step Abstraction}. For each action $\act_{i,t}$ within a segment, we generate an abstract representation by deriving a \emph{semantic element description} from the target element's attributes such as its accessibility label or surrounding text. This step explicitly discards the specific `raw element identifier' used in $\tau_i$, creating a transferable step node.

Finally, we compute embeddings for all new nodes and update the index to support efficient retrieval. This process is summarized in Algorithm~\ref{alg:construction}.

\begin{algorithm}[t]
\caption{Memory Construction Pipeline}
\label{alg:construction}
\small
\begin{algorithmic}[1]
\Require Successful trajectories $\mathcal{D} = \{(\task_i, \tau_i)\}$
\Ensure Hierarchical Memory Tree $\memtree$
\State Initialize $\memtree \leftarrow \emptyset$
\For{each trajectory $(\task_i,\tau_i)$ in $\mathcal{D}$}
  \Statex \textcolor{gray}{\textit{\# Level 1: Instruction Normalization}}
  \State $(I_i, C_i) \leftarrow \mathrm{Normalize}(\task_i)$
  \State $v^{task} \leftarrow \mathrm{GetOrCreateNode}(\memtree, I_i, C_i)$
  
  \Statex \textcolor{gray}{\textit{\# Level 2: Subgoal Segmentation}}
  \State $\{G_{i,k}\} \leftarrow \mathrm{Segment}(\tau_i)$
  \While{$\neg \mathrm{ConsistencyCheck}(\{G_{i,k}\})$}
      \State $\{G_{i,k}\} \leftarrow \mathrm{MergeSegments}(\{G_{i,k}\})$ 
  \EndWhile
  
  \For{each segment $G_{i,k}$}
    \State $(name, conds) \leftarrow \mathrm{Describe}(G_{i,k})$
    \State $v^{sub} \leftarrow \mathrm{CreateNode}(v^{task}, name, conds)$
    
    \Statex \textcolor{gray}{\textit{\# Level 3: Step Abstraction}}
    \For{each step $(\obs_{i,t}, \act_{i,t})$ in $G_{i,k}$}
      \State $(p_{i,t}, desc_{i,t}) \leftarrow \mathrm{AbstractStep}(\obs_{i,t}, \act_{i,t})$
      \State $\mathrm{CreateLeafNode}(v^{sub}, p_{i,t}, desc_{i,t})$
    \EndFor
  \EndFor
\EndFor
\State \textbf{return} $\memtree$
\end{algorithmic}
\end{algorithm}

\subsubsection{Induction Settings}

This pipeline supports two distinct induction settings. In the offline setting, we populate $\memtree$ once using a pre-existing dataset of successful trajectories such as the training split of Mind2Web. In the online setting, the agent starts with an empty memory and incrementally builds it. As it performs tasks in the evaluation stream, we monitor the success status of each episode; if an episode is successful, we immediately apply the abstraction pipeline to that trajectory and insert the new nodes into $\memtree$, allowing the agent to accumulate procedural knowledge during deployment.

\subsection{Stage-Aware Inference}
\label{sec:inference}

At test time, the agent determines the next action through a unified \emph{Stage-Aware Inference} process, as summarized in Algorithm~\ref{alg:online}. This process begins with a top-down retrieval strategy. First, we perform \emph{Task Retrieval} by querying $\memtree$ with the current instruction $\task$ to isolate workflows relevant to the user's high-level intent, yielding the top-$K_T$ task nodes $\mathcal{V}_T$. Next, for \emph{Subgoal Retrieval}, we query the subgoals attached to these tasks. The query combines the task instruction $\task$ with a summary of the history $\hist_{t-1}$ (recent actions) and an observation $\obs_t$ (salient elements). To prevent workflow mismatch caused by retrieving temporally distinct steps, we align the current page state with the appropriate workflow stage. We compute a combined score for each candidate subgoal $g$:
\begin{equation}
\resizebox{0.9\hsize}{!}{$
    \mathrm{Score}(g) = (1-\lambda) \cos(\mathbf{q}_G, \mathrm{emb}(g)) + \lambda \cdot \mathrm{CondMatch}(g, \obs_t),
$}
\label{eq:score}
\end{equation}
where $\mathbf{q}_G$ is the vector embedding of the combined query string, $\mathrm{emb}(g)$ denotes the embedding of the candidate subgoal $g$, $\lambda \in [0,1]$ is a hyperparameter balancing semantic similarity and condition matching, and $\mathrm{CondMatch}$ measures the lexical overlap (Jaccard similarity) between the subgoal's pre-conditions/post-conditions and the current observation summary.

\begin{algorithm}[t]
\caption{Stage-Aware Inference}
\label{alg:online}
\small
\begin{algorithmic}[1]
\Require Instruction $\task$, History $\hist_{t-1}$, Obs $\obs_t$, Memory $\memtree$
\Ensure Next action $\act_t$
\Statex \textcolor{gray}{\textit{\# Hierarchical Retrieval}}
\State $\mathcal{V}_T \leftarrow \mathrm{RetrieveTasks}(\memtree, \task)$
\State $\mathcal{G} \leftarrow \mathrm{RetrieveSubgoals}(\mathcal{V}_T, \hist_{t-1}, \obs_t)$

\Statex \textcolor{gray}{\textit{\# Planner: State Abstraction \& Stage Selection}}
\State $s_t \leftarrow \mathrm{AbstractState}(\obs_t)$
\State $(\hat{\sg}_t, c_t) \leftarrow \pi_{\text{plan}}(\task, s_t, \mathcal{G})$
\If{$\mathrm{IsLowConfidence}(\mathcal{G}, \hat{\sg}_t, c_t)$}
  \State $\mathcal{G}' \leftarrow \mathrm{ExpandSearch}(\mathcal{G})$
  \State $(\hat{\sg}_t, c_t) \leftarrow \pi_{\text{plan}}(\task, s_t, \mathcal{G}')$
  \If{$\mathrm{IsLowConfidence}(\mathcal{G}', \hat{\sg}_t, c_t)$}
     \State \textbf{return} $\pi_{\text{base}}(\task, \hist_{t-1}, \obs_t)$ 
  \EndIf
\EndIf

\Statex \textcolor{gray}{\textit{\# Actor: Action Grounding}}
\State $\mathcal{S} \leftarrow \mathrm{RetrieveSteps}(\hat{\sg}_t, \obs_t)$
\State $\act_t \leftarrow \pi_{\text{act}}(\task, \obs_t, \hat{\sg}_t, \mathcal{S}, \mathcal{E}_t)$
\State \textbf{return} $\act_t$
\end{algorithmic}
\end{algorithm}

Based on the retrieved candidates, we employ a \emph{Planner} and an \emph{Actor} to perform the following actions.
The \emph{Planner} performs a state abstraction and verification process to ensure temporal consistency. It abstracts the raw observation $\obs_t$ to identify the current logical stage by matching it with the pre-conditions and post-conditions of the retrieved subgoals in $\mathcal{G}$. This stage-aware selection effectively filters out invalid future or past steps, outputting the selected subgoal index $\hat{\sg}_t$ and a confidence score $c_t$, where $c_t$ represents the predicted probability $P(\hat{\sg}_t) \in [0,1]$ that the selected stage aligns with the current state.

To mitigate error propagation, we implement a \emph{Confidence-Aware Fallback} mechanism. We compute a robustness metric $\Delta = P(\hat{\sg}_{top1}) - P(\hat{\sg}_{top2})$ as the probability margin between the top-1 and top-2 ranked subgoals. If $\Delta$ falls below a margin threshold $\delta$ or if the absolute confidence $c_t$ is lower than a threshold $\tau$, we declare a low-confidence state. In such cases, the agent broadens the retrieval scope by increasing the number of candidate subgoals $K_G$. If uncertainty persists after expansion, the agent reverts to a baseline policy without memory conditioning, preventing it from being misled by irrelevant retrieval.

Once a valid stage is confirmed, we retrieve the top-$K_S$ step nodes based on similarity to the current observation. These nodes serve as few-shot exemplars $\mathcal{S}$, providing the Actor with action patterns and semantic element descriptions relevant to the current stage. Then, the Actor generates the concrete action. It receives the instruction, observation, selected subgoal, and retrieved step exemplars $\mathcal{S}$. Crucially, the Actor does not directly copy IDs from the exemplars. Instead, it uses the retrieved \emph{semantic element descriptions} to locate the corresponding element in the current candidate set $\mathcal{E}_t$. This grounding process involves querying the LLM to select the element $e \in \mathcal{E}_t$ that best matches the semantic description, such as ``button labeled `Search' '', rather than the raw identifier. This ensures the action is valid on the current page even if the underlying DOM IDs have changed.

\subsection{Generalization}
\label{sec:generalization}

While presented in a text-based DOM context, \hmt is adaptable to other modalities and environments.
For environments utilizing screenshots, the semantic element description in step nodes can be augmented with visual embeddings of the element's cropped region, allowing retrieval to operate in a joint text-visual embedding space.
Furthermore, since different benchmarks require different action formats, such as IDs versus coordinates, \hmt maintains memory in an abstract form. A lightweight \emph{resolver} module maps the Actor's semantic output to the specific format required by the environment. For ID-based environments like WebArena, it selects the element maximizing the match with the predicted description, and for coordinate-based environments, it invokes a grounding model to predict $(x, y)$ coordinates.

\begin{table*}[tp]
\centering
\caption{Mind2Web results. The best results are \textbf{bolded}, and the second-best are \underline{underlined}. \hmt shows distinct advantages in cross-website generalization.}
\label{tab:mind2web_main}
\begin{adjustbox}{max width=\textwidth}
\begin{tabular}{lcccc|cccc|cccc}
\toprule
& \multicolumn{4}{c|}{Cross-Task} & \multicolumn{4}{c|}{Cross-Website} & \multicolumn{4}{c}{Cross-Domain} \\
Method & EA & AF1 & StepSR & TaskSR & EA & AF1 & StepSR & TaskSR & EA & AF1 & StepSR & TaskSR \\
\midrule
MindAct~\cite{deng2023mind2web} & 66 & \underline{60.6} & 36.2 & 2.0 & 35.8 & \underline{51.1} & 30.1 & 2.0 & 21.6 & \textbf{52.8} & 26.4 & \textbf{2.0} \\
AWM$_{\text{offline}}$~\cite{wang2024agent} & \underline{50.6} & 57.3 & \underline{45.1} & \textbf{4.8} & 41.4 & 46.2 & 33.7 & \underline{2.3} & 36.4 & 41.6 & 32.6 & 0.7 \\
AWM$_{\text{online}}$~\cite{wang2024agent}  & 50.0 & 56.4 & 43.6 & 4.0 & \underline{42.1} & 45.1 & \underline{33.9} & 1.6 & \underline{40.9} & 46.3 & \underline{35.5} & 1.7 \\
\midrule
\emph{\hmt (Ours)} & \textbf{54.2} & \textbf{63.5} & \textbf{48.5} & \underline{4.6} & \textbf{48.8} & \textbf{54.5} & \textbf{39.7} & \textbf{3.2} & \textbf{42.3} & \underline{48.8} & \textbf{37.1} & \underline{1.9} \\
\bottomrule
\end{tabular}
\end{adjustbox}
\end{table*}

\begin{table*}[tp]
\centering
\caption{WebArena results. HMT improves performance on logic-heavy domains (GitLab, CMS), but shows limitations in spatial-heavy domains (Maps). The total TaskSR is a weighted average. Best in \textbf{BOLD}, second best \underline{UNDERLINED}.}
\label{tab:webarena_main}
\begin{adjustbox}{max width=\textwidth}
\begin{tabular}{lc|ccccc|c} 
\toprule
Method & Total TaskSR & Shopping & CMS & Reddit & GitLab & Maps & \#Steps (Avg) \\
\midrule
WebArena baseline [48] & 14.9 & 14.0 & 11.0 & 6.0 & 15.0 & 16.0 & 12.4 \\
SteP [49] & 33.0 & 37.0 & 24.0 & 59.0 & 32.0 & 30.0 & 8.5 \\
AutoEval [50] & 20.2 & 25.5 & 18.1 & 25.4 & 28.6 & 31.9 & 46.7 \\
AWM [5] & \underline{35.5} & 30.8 & 29.1 & 50.9 & 31.8 & 43.3 & \underline{5.9} \\
\midrule
Flat Retrieval (Baseline) & 32.1 & 28.5 & 25.8 & 47.0 & 28.5 & 39.0 & 6.5 \\
HMT (\textit{Ours}) & \textbf{38.7} & 33.8 & 34.1 & 52.5 & 37.6 & 42.2 & \textbf{5.2} \\
\bottomrule
\end{tabular}
\end{adjustbox}
\end{table*}

\section{Experiments}
\label{sec:experiments}

This section evaluates \hmt under two distinct memory-induction settings. First, we examine an offline setting where memory is constructed from a pre-existing corpus. Second, we explore an online setting where memory is incrementally built during deployment.

\subsection{Settings}

\subsubsection{Benchmarks and Datasets}
\label{sec:benchmarks}

We utilize \emph{Mind2Web}~\cite{deng2023mind2web} to evaluate generalization capabilities in the offline setting. This benchmark uses a large-scale dataset containing over 2,000 open-ended web tasks collected from 137 real-world websites. Following the original protocol~\cite{deng2023mind2web}, we evaluate on three distinct test splits to assess different levels of generalization:
\begin{itemize}
    \item \emph{Cross-Task}: This split contains unseen tasks from websites that were seen during training. In this setting, the agent has likely encountered similar page layouts and DOM structures.
    \item \emph{Cross-Website}: This split includes tasks from websites that were never seen during training, although the domains (e.g., Travel, Shopping) are familiar. This setup assesses the agent's ability to adapt to entirely new website structures within a known task context.
    \item \emph{Cross-Domain}: This split holds out entire top-level domains (e.g., Information, Service) from training. The agent is expected to generalize to completely new domains without prior exposure to their specific websites or task logic.
\end{itemize}

To assess interactive execution in the online setting, we employ \emph{WebArena}~\cite{zhou2024webarena}. This environment spans multiple domains including Shopping, CMS, GitLab, and Maps, requiring the agent to navigate multi-turn tasks to achieve a goal. Success is determined by execution-based validators.

\subsubsection{Comparison Methods}
\label{sec:baselines}

We compare our proposed \hmt with several state-of-the-art methods. These methods include general-purpose agents (\emph{MindAct}~\cite{deng2023mind2web} and \emph{WebArena Baseline}~\cite{zhou2024webarena}) and hierarchical or retrieval-augmented agents (\emph{SteP}~\cite{sodhi2023step}, \emph{AutoEval}~\cite{pan2024autoeval}, and \emph{Agent Workflow Memory (AWM)}~\cite{wang2024agent}). \emph{AWM} induces linear workflows from successful trials and serves as a direct comparison for our memory-based approach. 

To rigorously evaluate the contribution of our hierarchical structure, we construct a degraded variant of our method as a baseline \textbf{Flat Retrieval}: This variant stores successful trajectories as a flat list of steps and retrieves the top-$K$ steps based on embedding similarity, without utilizing the proposed stage-aware planning or semantic element descriptions.

\subsubsection{Implementation Details}
\label{sec:impl_details}

We use GPT-4 as the backbone model for all components. The decoding temperature is set to zero. Based on preliminary validation, we set the retrieval widths for tasks, subgoals, and step exemplars to $K_T = 5$, $K_G = 8$, and $K_S = 5$, respectively. The interaction history is truncated to $N_h = 6$ actions, and the observation is summarized to the top $N_e = 30$ salient elements. In Eq. (3), the weighting factor is $\lambda = 0.3$. The fallback thresholds are set to $\delta = 0.1$ and $\tau = 0.15$.

\subsection{Main Results}
\label{sec:main_results}

The comparison results on Mind2Web and WebArena are shown in Table~\ref{tab:mind2web_main} and Table~\ref{tab:webarena_main}, respectively. From the results, we make several observations.

First, on the \emph{Mind2Web} benchmark (Table~\ref{tab:mind2web_main}), \hmt achieves superior performance across all splits. In the \emph{Cross-Task} split, \hmt performs comparably to \emph{AWM} (48.5 vs. 45.1 in StepSR and 54.2 vs. 50.6 in EA), indicating that when the DOM structure remains static, flat memory with raw identifiers is sufficient. However, in the critical \emph{Cross-Website} split, \hmt significantly outperforms existing methods, improving StepSR by 6.0\% compared to \emph{AWM}. This validates that when site-specific IDs fail due to layout changes, our semantic element descriptions successfully bridge the grounding gap.

Second, on the \emph{WebArena} benchmark (Table~\ref{tab:webarena_main}), \hmt achieves the highest total success rate of 38.7\%. We observe substantial improvements in logically complex domains like \emph{GitLab} (+5.8\%) and \emph{CMS} (+5.0\%). This suggests that the Intent-Stage hierarchy effectively prevents the agent from losing track of the workflow in long-horizon tasks.

Third, regarding the \emph{Maps} domain in WebArena, \hmt performs slightly worse than \emph{AWM} (42.2\% vs. 43.3\%). We attribute this to the fact that map interactions often rely on spatial coordinates, where our semantic text-based abstraction offers less benefit than raw trajectory replay. Furthermore, the average number of steps is reduced (5.2 vs. 5.9), indicating that hierarchical planning reduces redundant exploration.

\begin{figure*}[t]
  \centering
  \includegraphics[width=0.95\textwidth]{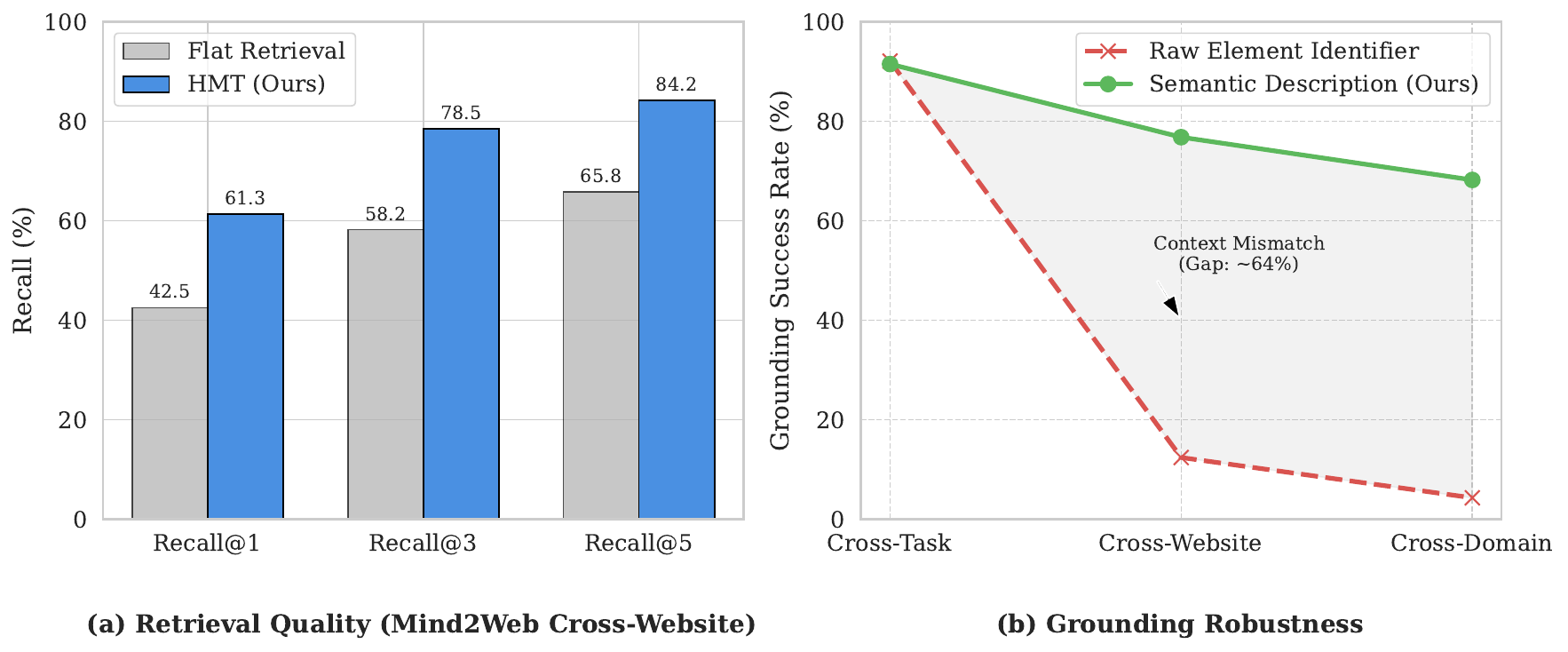}
  \caption{Mechanism Analysis. (a) Retrieval recall comparisons show that \hmt provides more accurate context. (b) Grounding success rate across generalization splits shows that raw identifiers fail in cross-website and cross-domain settings, while the semantic descriptions (ours) remain robust.}
  \label{fig:mechanism}
\end{figure*}

\begin{figure*}[t]
  \centering
  \includegraphics[width=\textwidth]{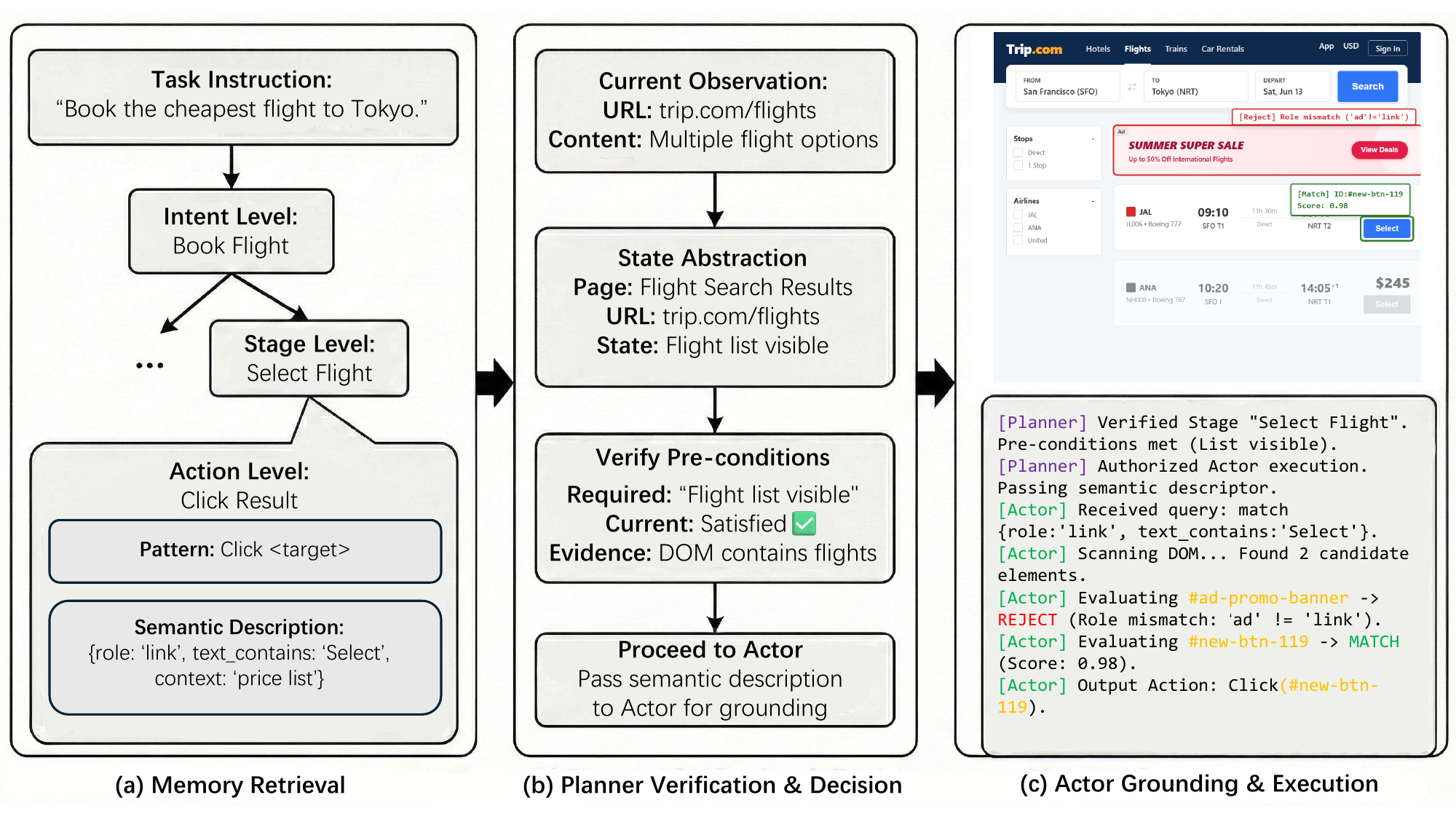}
  \caption{Visual analysis of a successful cross-website grounding trace demonstrated by \hmt. 
  \textbf{(a) Memory Retrieval:} The agent retrieves an abstract action pattern and a semantic descriptor from the hierarchical memory, explicitly discarding the site-specific raw identifier (\texttt{\#btn-sfo-136}) from the source trace.
  \textbf{(b) Planner Verification \& Decision:} To prevent \textbf{workflow mismatch}, the Planner verifies that the current page satisfies the stage pre-conditions (e.g., ``Flight list visible'') before authorizing the Actor, ensuring actions are only executed in the correct context.
  \textbf{(c) Actor Grounding \& Execution:} Guided by the semantic descriptor, the Actor scans the target DOM on \textit{Trip.com}. It successfully distinguishes between a distractor advertisement and the correct flight selection button, executing the correct action despite the layout shift.}
  \label{fig:vis_success}
\end{figure*}

\begin{figure*}[t]
  \centering
  \includegraphics[width=\textwidth]{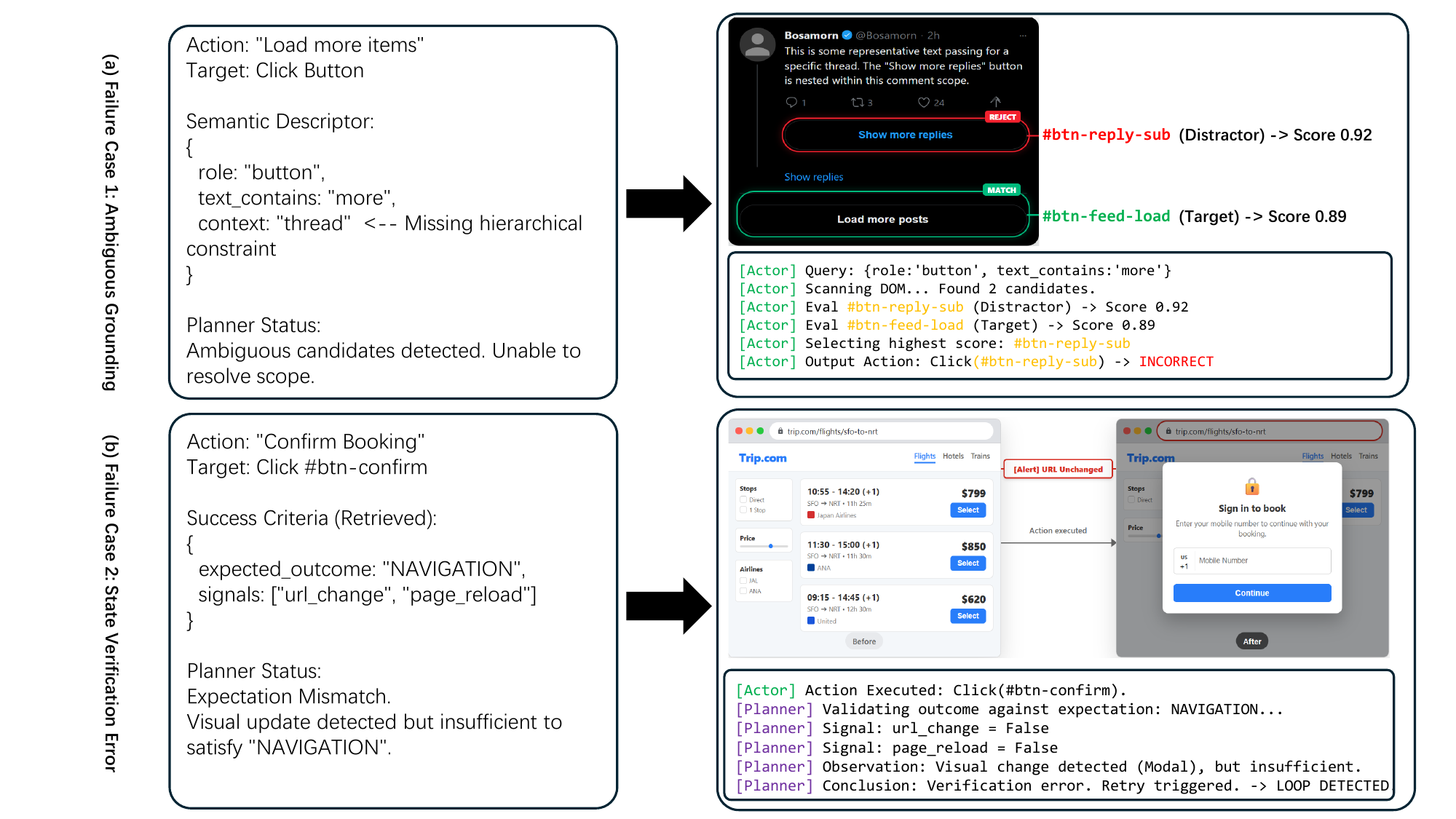}
  \caption{Analysis of two representative failure modes in cross-website grounding. 
  \textbf{(a) Ambiguous Grounding:} The agent retrieves a generic descriptor (\texttt{text\_contains: more}) lacking hierarchical constraints. Consequently, the Actor assigns a higher similarity (score 0.92) to the distractor candidate than to the intended target (score 0.89), leading to an execution error.
  \textbf{(b) State Verification Error:} The Planner retrieves a strict success criterion expecting a \texttt{NAVIGATION} event (e.g., URL change or page reload). Although the action triggers a visual update (modal popup), the URL remains static. The Planner's rigid verification logic (\texttt{url\_change = False} and \texttt{page\_reload = False}) fails to recognize the visual progress, incorrectly marking the step as incomplete and initiating a retry loop.}
  \label{fig:vis_failure}
\end{figure*}

\subsection{Ablation Studies}
\label{sec:ablations}

To evaluate the effectiveness of individual components, we design four variants of our method for comparison: 
(1) \textbf{w/ Flat Memory}: we remove the task and subgoal hierarchy to retrieve steps directly from a flat pool, which is equivalent to the \textit{Flat Retrieval} baseline. This variant evaluates the importance of the hierarchical structure;
(2) \textbf{w/o Pre/Post-conditions}: we remove the explicit pre-conditions and post-conditions, causing the Planner to rely solely on textual similarity for stage selection. This variant evaluates the effectiveness of state-aware planning;
(3) \textbf{w/ Raw Element Identifiers}: we replace the semantic element descriptions with raw element identifiers extracted from the source website, in order to evaluate the effectiveness of semantic grounding;
(4) \textbf{w/o Planner}: we remove the explicit Planner to test the single Actor setup where the model implicitly selects steps;
(5) \textbf{w/o Confidence Fallback}: we disable the confidence check to assess the importance of the fallback mechanism.

The results of ablation studies are shown in Table~\ref{tab:ablation}. We have the following observations: 
(1) Our method substantially outperforms ``w/ Flat Memory'', which indicates that the hierarchical structure reduces context pollution in long-horizon tasks. For example, a gain of 6.6\% is achieved on WebArena.
(2) With the removal of pre/post-conditions, the performance drops by 2.5\% on WebArena. This demonstrates the importance of considering observable state changes for stage alignment.
(3) Most critically, replacing semantic descriptions with raw identifiers (``w/ Raw Element Identifiers'') proves catastrophic for cross-website generalization. On the Mind2Web Cross-Website split, the StepSR plummets from 39.7\% to 12.4\%, verifying that raw identifiers are non-transferable when the DOM structure changes.
(4) Removing the Planner (``w/o Planner'') degrades performance, highlighting the necessity of decomposing planning and execution.

\begin{table}[t]
\centering
\caption{Ablation study results. We measure the impact of key components on Mind2Web StepSR and WebArena TaskSR.}
\label{tab:ablation}
\begin{adjustbox}{max width=\linewidth}
\begin{tabular}{lcc}
\toprule
Method Variant & \shortstack{Mind2Web (Cross-Website)\\StepSR (\%)} & \shortstack{WebArena\\Total TaskSR (\%)} \\
\midrule
\textbf{\emph{Full \hmt}} & \textbf{39.7} & \textbf{38.7} \\
\midrule
w/ Flat Memory & 33.2 & 32.1 \\
w/o Pre/Post-conditions & 37.1 & 36.2 \\
w/ Raw Element Identifiers & 12.4 & 34.5 \\
w/o Planner & 35.8 & 33.5 \\
w/o Confidence Fallback & 38.9 & 37.8 \\
\bottomrule
\end{tabular}
\end{adjustbox}
\end{table}

\subsection{Mechanism Analysis}
\label{sec:mechanism}

To provide deeper insights into why \hmt works, we conduct focused quantitative analyses regarding retrieval quality and grounding robustness.

\subsubsection{Retrieval Quality}
We assess whether the hierarchical structure improves the relevance of retrieved memories by measuring the Recall@5 of the ground-truth action step on the Mind2Web \emph{Cross-Website} split. As shown in Fig.~\ref{fig:mechanism}(a), \hmt achieves a recall of 84.2\%, significantly outperforming the 65.8\% obtained by the flat retrieval baseline. This improvement indicates that the stage-level constraints effectively filter out temporally irrelevant noise, ensuring the agent focuses on contextually valid actions.

To illustrate this difference concretely, Table~\ref{tab:retrieval_case} presents a real-world retrieval case on the \emph{TripAdvisor} search result page. The user intent is to ``Find a flight.'' The flat retrieval baseline is misled by semantic overlap, retrieving checkout actions from a previous \emph{Expedia} task. In contrast, the \hmt Planner aligns the state with the ``Browse \& Select'' stage, ensuring the top candidates are all functionally relevant to the current page.

\begin{table*}[tp]
\centering
\caption{Comparison of top-3 retrieved memories on TripAdvisor. The task is ``Find a flight.'' Flat retrieval fetches temporally misaligned actions (e.g., Checkout) due to semantic overlap, while \hmt restricts retrieval to the current workflow stage.}
\label{tab:retrieval_case}
\small
\begin{adjustbox}{max width=\textwidth}
\begin{tabular}{c|l|l|c|l}
\toprule
\textbf{Rank} & \textbf{Method} & \textbf{Retrieved Action Content (Simplified)} & \textbf{Valid?} & \textbf{Reasoning} \\
\midrule
\multirow{3}{*}{\shortstack{Flat\\Retrieval}} 
& 1 & Action: Click ``Place Order'' (from Expedia Checkout) & \textcolor{red}{\ding{55}} & Stage Mismatch: Attempting to checkout before selecting item. \\
& 2 & Action: Click ``Add to Cart'' (from Expedia Detail Page) & \textcolor{red}{\ding{55}} & Context Pollution: Element does not exist on search page. \\
& 3 & Action: Click Product Title (from Expedia Search) & \textcolor{green}{\ding{51}} & Correct action, but ranked low due to noise. \\
\midrule
\multirow{3}{*}{\shortstack{\hmt\\(Ours)}} 
& 1 & \textbf{Pattern}: Click Item \quad \textbf{Descriptor}: \{Role: link, Text: match query\} & \textcolor{green}{\ding{51}} & Stage Aligned: Correctly identifies browsing behavior. \\
& 2 & \textbf{Pattern}: Sort Price \quad \textbf{Descriptor}: \{Role: dropdown, Text: 'Sort'\} & \textcolor{green}{\ding{51}} & Valid Option: Contextually relevant for price constraints. \\
& 3 & \textbf{Pattern}: Next Page \quad \textbf{Descriptor}: \{Role: button, Text: 'Next'\} & \textcolor{green}{\ding{51}} & Valid Option: Standard navigation on search pages. \\
\bottomrule
\end{tabular}
\end{adjustbox}
\end{table*}

\subsubsection{Grounding Robustness}
We further quantify the transferability of our action abstraction by comparing the success rate of locating the target element using raw identifiers versus our semantic descriptions. As illustrated in Fig.~\ref{fig:mechanism}(b), while raw identifiers perform acceptably in cross-task settings (92.1\%), their performance collapses to 12.4\% in cross-website settings. In contrast, semantic descriptors maintain a robust success rate of 76.8\%. This contrast confirms that semantic matching is the key factor in resolving intention-execution entanglement.

\subsection{Efficiency Analysis}
\label{sec:efficiency}

Beyond accuracy, we evaluate the computational efficiency of \hmt. Table~\ref{tab:efficiency} compares the average token consumption and latency per step on \emph{WebArena}. By abstracting raw HTML trajectories into compact semantic descriptions, \hmt reduces the average context length by approximately 72.7\% compared to standard retrieval. Despite the two-stage inference process (Planner then Actor), the significant reduction in input tokens results in a lower overall latency (3.5s vs. 5.2s) and a 71.0\% reduction in inference cost per task.

\begin{table}[h]
\centering
\caption{Efficiency comparison on WebArena. \hmt reduces token consumption and latency by compressing raw trajectories into semantic nodes.}
\label{tab:efficiency}
\begin{adjustbox}{max width=\linewidth}
\begin{tabular}{lccc}
\toprule
Method & \shortstack{Avg. Context\\Tokens} & \shortstack{Latency\\(s/step)} & \shortstack{Cost\\(\$/Task)} \\
\midrule
Flat Retrieval (Raw HTML) & 11,450 & 5.2 & 0.38 \\
\textbf{\emph{\hmt (Ours)}} & \textbf{3,120} & \textbf{3.5} & \textbf{0.11} \\
\bottomrule
\end{tabular}
\end{adjustbox}
\end{table}

\subsection{Qualitative Analysis}
\label{sec:qualitative}

To provide a granular understanding of the proposed framework, we conduct a qualitative trace analysis of the decision-making process in both success and failure scenarios. This analysis highlights how the hierarchical structure facilitates knowledge transfer and identifies boundary conditions where current abstractions may fall short.

We first visualize a successful cross-website knowledge transfer trace in Fig.~\ref{fig:vis_success}, where the agent is tasked with booking a flight on \textit{Trip.com} using memory derived from \textit{Expedia}. The process begins with the retrieval of a relevant action node. Crucially, the memory construction phase has already discarded the site-specific raw identifier from the source website, retaining only a transferable semantic description consisting of the element's role and textual content. Upon retrieval, the Planner initiates the inference process. It performs a state abstraction on the raw page observation and confirms that the pre-condition, specifically the visibility of a flight list, is satisfied. This step prevents the execution of actions in incorrect contexts. Subsequently, the Planner delegates execution to the Actor, passing down the semantic description. The Actor then scans the current DOM tree for matching elements. As illustrated in the execution log, the Actor encounters a distractor element, an advertisement with an ID of \texttt{\#ad-promo-banner}, which semantically resembles the target but possesses a conflicting role attribute. The Actor correctly rejects this candidate based on the role mismatch and successfully grounds the action to the correct flight selection button identified as \texttt{\#new-btn-119}, which exhibits a high semantic match score. This example demonstrates how the decomposition of planning and grounding effectively resolves the intention-execution entanglement.

Despite these capabilities, the system exhibits limitations in scenarios involving Ambiguous Grounding or subtle state transitions, as detailed in Fig.~\ref{fig:vis_failure}. The first failure mode, Ambiguous Grounding, occurs when the retrieved semantic description lacks sufficient hierarchical context. In the social media example shown in Fig.~\ref{fig:vis_failure}(a), the agent intends to load more posts at the bottom of the feed. However, the interface contains a distractor button ``Show more replies'' within a specific comment thread. Since both buttons match the generic description of containing the text ``more'', the Actor identifies multiple high-scoring candidates. Lacking a precise parent-child structural constraint in the memory, the Actor assigns a higher similarity score to the distractor candidate than to the intended target ``Load more posts'', leading to an execution error. The second failure mode, State Verification Error, is observed in Single Page Applications (SPAs) where visual updates do not trigger URL changes. As shown in Fig.~\ref{fig:vis_failure}(b), the agent successfully clicks the confirmation button, triggering a modal popup. However, the Planner's rigorous post-condition check requires a URL transition or a page reload to confirm stage completion. Since the URL remains static, the Planner incorrectly perceives the action as failed and initiates a retry loop, repeatedly executing the same action despite the visual progress. These cases suggest that future iterations of hierarchical memory should incorporate richer structural context descriptors and more versatile state verification mechanisms beyond DOM and URL signals.

\section{Conclusion}

In this work, we focus on the challenge of cross-website generalization by proposing the Hierarchical Memory Tree (\hmt), a structured framework that explicitly decouples logical planning from site-specific action execution. By autonomously constructing a hierarchy of standardized intents, stage-aware subgoals defined by observable pre-conditions and post-conditions, and semantic element descriptions, \hmt effectively mitigates the intention-execution entanglement inherent in flat memory systems. Our stage-aware inference mechanism, comprising a Planner and an Actor, not only ensures robust grounding in novel environments but also significantly improves computational efficiency by compressing raw trajectories. Extensive evaluations on Mind2Web and WebArena confirm that \hmt establishes a new standard for transferable web agents, paving the way for more scalable and autonomous systems capable of lifelong learning.

\bibliographystyle{IEEEtran}
\bibliography{references}

\end{document}